\documentclass[12pt]{iopart}

\usepackage{graphicx} 
\usepackage{cite}
\usepackage[version=4]{mhchem}
\usepackage{graphicx}
\usepackage{framed} 
\usepackage{amssymb,amsfonts}
\usepackage{cleveref}
\usepackage{caption}
\usepackage{subcaption}
\usepackage[table,xcdraw]{xcolor}
\usepackage{moresize}
\usepackage[normalem]{ulem}
\usepackage[hyphens]{url}
\usepackage{breakurl}
\usepackage{comment}

\usepackage{algorithm}
\usepackage{algorithmic} 

\graphicspath{{figures/}}

\begin{document}

\title{An energy-efficient spiking neural network with continuous learning for self-adaptive brain-machine interface}
\author{Zhou Biyan,
	and~Arindam~Basu$^{*}$}
\address{City University of Hong Kong, Hong Kong}

\address{$^{*}$ Correspondence author}
\ead{arinbasu@cityu.edu.hk}
\date{xx/xx/2025}
\maketitle

\begin{abstract}

The number of simultaneously recorded neurons follows an exponentially increasing trend in implantable brain-machine interfaces (iBMIs). Integrating the neural decoder in the implant is an effective data compression method for future wireless iBMIs. However, the non-stationarity of the system makes the performance of the decoder unreliable. To avoid frequent retraining of the decoder and to ensure the safety and comfort of the iBMI user, continuous learning is essential for real-life applications. Since Deep Spiking Neural Networks (DSNNs) are being recognized as a promising approach for developing resource-efficient neural decoder, we propose continuous learning approaches with Reinforcement Learning (RL) algorithms adapted for DSNNs. Banditron and AGREL are chosen as the two candidate RL algorithms since they can be trained with limited computational resources, effectively addressing the non-stationary problem and fitting the energy constraints of implantable devices. To assess the effectiveness of the proposed methods, we conducted both open-loop and closed-loop experiments. The accuracy of open-loop experiments conducted with DSNN\_Banditron and DSNN\_AGREL remains stable over extended periods. Meanwhile, the time-to-target in the closed-loop experiment with perturbations, DSNN\_Banditron performed comparably to that of DSNN\_AGREL while achieving reductions of 98\% in memory access usage and 99\% in the requirements for multiply-and-accumulate (MAC) operations during training. Compared to previous continuous learning SNN decoders, DSNN\_Banditron requires 98\% less computes making it a prime candidate for future wireless iBMI systems.
 
\end{abstract}

\begin{flushleft}\label{AbbreviationsList}
\footnotesize{
    \textbf{List of Abbreviations- } 
    
    \begin{table}[h]
        \begin{tabular}{ll}
            iBMI    & Implantable Brain Machine Interface\\
            NHP    & Non-human Primate\\
            SNN    & Spiking Neural Network\\
            ANN     & Artificial Neural Network \\
            RL   & Reinforcement learning 
        \end{tabular}
    \end{table}
}
\end{flushleft}

\section{Introduction}
\label{sec:introduction} 

\begin{figure}[!t]
\begin{framed}
    \centering
    \includegraphics[width=0.88\textwidth]{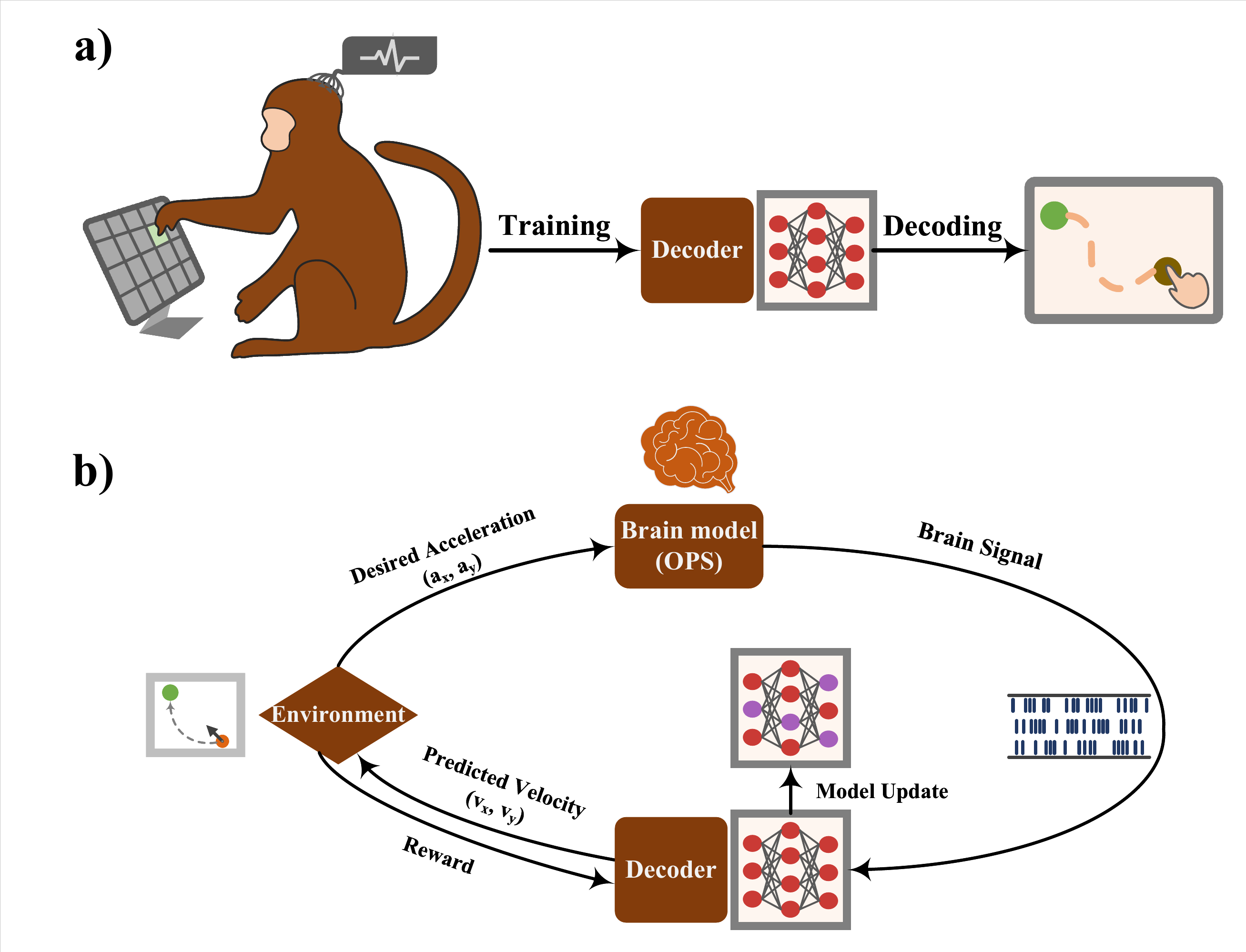}
    \caption{(a) Schematic illustration of an open-loop experiment where decoders are trained using previously obtained brain signals and tested. (b) Flow diagram of a closed-loop experiment that includes an environment for participants to operate, a brain model simulator (OPS), and a trainable decoder.}
    \label{fig:openclosedloop}
\end{framed}
\end{figure}

Implantable Brain-Machine Interface (iBMI) depicted in \cref{fig:openclosedloop} (a) creates a connection between the brain and machines, serving as a promising technology for interpreting a patient's intent to control robotic arms \cite{bmi_arm}, computer cursors \cite{pandarinath2017high}, or an actuator\cite{milin_nature}. Besides, this technology is specifically developed as an advanced assistive tool aimed at enabling numerous individuals with paralysis \cite{CRF} to recover their ability to engage in daily living activities \cite{camilo_plos}, restore naturalistic spoken communications \cite{littlejohn2025streaming}, or modulate voice in real-time \cite{wairagkar2025instantaneous}. 

However, the majority of BMI systems rely on a wired connection to a computer, which introduces several concerns: the risk of infection at the opening of the skull and limitations on the range of motion.  Studies indicate that users consider independence to be their primary concern \cite{collinger2013functional}, highlighting the potential of wireless iBMI as an effective solution \cite{nurmikko_wireless}. The scalability of wireless iBMI presents a significant challenge due to its limited bandwidth of Mbps \cite{borton2013implantable} (limited to recording about 150-200 neurons). On the other hand, there is a notable trend toward an exponential increase in the number of simultaneously recorded neurons, which is projected to reach more than 1000 \cite{stevenson_2013}. Moreover, the power dissipation needs to be limited for cortical implants at 80 mW/cm$^{2}$ to ensure patient safety \cite{kim2006preliminary}. Thus, there is a conflict between the requirements of high data rate and low power consumption. The compression of neural data is crucial to meet the energy consumption and data transmission rate constraints associated with neural implants. There are various compression methods available, which include spike detection \cite{chae2009128}, spike sorting \cite{basu2017big}, and decoding \cite{shaikh2019towards}. Notably, decoding provides the highest level of compression among these techniques \cite{shaikh2019towards}. It also helps in preserving data privacy since the raw neural data does not have to be transmitted. In order to meet the requirements for implants, it is essential for the decoder to achieve an optimal balance between accuracy and resource utilization. 

One of the traditional decoders is the Kalman Filter \cite{an2022power}, whereas artificial neural networks (ANNs) are increasingly used to analyze nonlinear and high-dimensional features \cite{shaeri202433}. Research indicates that neural activity can adapt and change over time \cite{sorrell2021brain}, resulting in non-stationarity in neural data that presents significant challenges for decoders with fixed weights. As a result, the recalibration of decoders is essential. However, traditional calibration methods require considerable amounts of data and may necessitate the suspension of equipment operation during the training process. This highlights the critical need for a more efficient approach to decoder calibration. 

There are three principal strategies that can be employed to reduce the impact of the nonstationarity of neural activities to restore the decoder performance. The first approach is transfer learning \cite{wu2020transfer}, which represents a domain adaptation strategy that transfers knowledge from the source domain to the target domain. By leveraging the knowledge acquired by pre-trained models from related domains, transfer learning effectively reduces the impact of discrepancies between the target domain and the source domain. A deep convolutional neural network (CNN) pre-trained on EEG data from multiple individuals effectively transferred its knowledge to new individuals through fine-tuning \cite{8008420}, significantly reducing calibration time while requiring minimal data. The second method employs an adaptive decoder that is capable of being trained on historical data and continuously calibrated with newly acquired information \cite{jarosiewicz2015virtual}. This approach allows for ongoing enhancement in the accuracy and effectiveness of decoders while eliminating the need for frequent recalibration of the decoders \cite{wilson2023long}. The third approach is to develop a powerful enough decoder with enhanced generalization capabilities to address the challenges of non-stationary neural signal \cite{sussillo2016making}. Combining the first and second methods above and considering the low power requirement of iBMI systems, our work adapts the concept of transfer learning, focusing on a low-complexity method that can continuously calibrate.

Compared to ANNs, spiking neural networks (SNNs) are more energy-efficient due to their event-driven nature \cite{liao2022energy, snn_review}. SNNs utilize discrete events known as spikes, allowing them to capture complex patterns and timing relationships in data through their temporal dynamics. A recent effort \cite{yik2025neurobench} presented a benchmark framework for quantifying neuromorphic approaches, and one chosen task is that of motor decoding. This highlights the significant potential for energy efficiency in SNNs. Although this task focuses on an open-loop system, a closed-loop system is more representative of real-life situations like \cref{fig:openclosedloop} (b); it is also understood that better closed-loop benchmarks are necessary for improvements in neuromorphic computing\cite{davies_benchmark}. In a closed-loop system, participants receive feedback from the iBMI's output, observe subsequent changes, and attempt to correct the output. In real life, the system is dynamic, and the decoder needs to self-adapt to maintain its performance. However, the design of most decoders does not consider the resource requirements during their updates. We make the following novel contributions in this paper:
\begin{itemize}
\item We adapt reinforcement learning (RL) algorithms (Banditron and AGREL) within SNNs to enable continuous adaptation and effectively address non-stationary challenges.
\item To apply Banditron to deep SNNs, we incorporate transfer learning in SNN decoders that also enables  an energy-efficient decoding mechanism.
\item We demonstrate the application of Banditron for velocity prediction tasks by recasting the regression problem in a classification framework.
\item Our method demonstrates better performance than untuned decoders and involves significantly less complexity in updates compared to methods that require changing all weights in a deep network.
\item We show both open-loop and closed-loop experiments to highlight the benefits of our approach.
\end{itemize}

\section{Related Works}
\label{sec:relatedworks}

The initial designs of decoders primarily concentrated on linear decoders, such as the well-known vector algorithm \cite{georgopoulos1986neuronal} and its generalized form, the Optimal Linear Estimators \cite{collinger2013high}. Other notable methods included Bayesian algorithms \cite{koyama2010comparison}, Winner filter \cite{kim2008neural}, Kalman filter and its variants \cite{murmann_nature}. These algorithms provide estimates of kinematics by utilizing a linear function that is based on binned spike counts. However, the linear decoder proved only to be optimal for linear variables and Gaussian noise \cite{makin2018superior}. Therefore, there is a growing interest in the application of machine learning decoders due to their capability to process nonlinear variables and handle complex data. Such as SVM decoder \cite{sharma2016using}\cite{friedenbergneuroprosthetic}, ANN-based decoder \cite{shaikh2019towards} \cite{willsey2022real} \cite{AutoEncoder}, RNN-based decoder \cite{speech_decode_nature2} \cite{Brain2Text} \cite{ZZ_FRM_SPD} \cite{littlejohn2025streaming}, and transformer-based decoder \cite{wairagkar2025instantaneous}. While these decoders achieve good performance, most of them are too complex for inclusion in an implant. Alternatively, neuro-inspired SNN-based decoders\cite{Liao2022,yik2025neurobench,imec_decoder} represent a promising trend in the design of BMI decoders, primarily due to their superior energy efficiency, which is a critical advantage for power-constrained iBMI. However, these earlier works all used fixed decoder designs (performance reported with daily retraining) limiting their usage over a long time due to the non-stationarity in neural activity. Also, they did not show any results for closed-loop experiments that are representative of real iBMI usage and are known to have different optimal parameters compared to open-loop ones\cite{cunningham2011closed}.  
 

To overcome the data non-stationarity problem in chronic implants, some decoders have attempted to integrate reinforcement learning (RL) methodologies to facilitate long-term adaptive learning in closed-loop settings. For a closed-loop BMI system, a control signal and a decoder are necessary. The participant and the BMI control agent are required to co-adapt in order to learn from the dynamic environment within the conventional closed-loop system \cite{taylor2003information}. Reinforcement learning (RL) is inspired by principles of operant conditioning observed in biological systems, where the learner discovers which actions yield the most reward through trial and error \cite{hilgard1966theories}. Some RL-based BMIs \cite{digiovanna2007brain,4540104} have aimed to develop a control strategy that utilizes the neuronal state of the user and the intended actions in tasks, without explicit guidance on which specific actions are deemed most appropriate. Also, research shows that biological sources contain simple binary scalar reward information, which can be used to train RL models \cite{prins2017feedback,benyamini2019neural}. Attention gating mechanisms\cite{yiwen_agrel} have also been used in RL based decoding to improve performance. However, all these works used traditional ANN which do not have the sparsity benefits of SNNs and hence have much higher computational complexity. A recent work \cite{taeckens2024spiking} has combined SNNs with a form of continual learning to tackle this issue. However, while the forward computation path benefits from sparsity, the continual learning algorithm does not use spikes and has a high computational cost for the update as we show in the Appendix. Another approach has been to use hybrid strategies combining ANN and event-driven nature of SNN for efficiency. For example, Event-based Gated Recurrent Unit (EGRU) \cite{evnn2023} utilized discrete events for communication between neurons, enhancing the efficiency of both the forward and backward paths. While this is an interesting approach, we show later how our proposed DSNN\_Banditron improves performance and computational efficiency over it.

In this study, we demonstrate that SNN combined with energy-efficient RL algorithms like Banditron\cite{shoeb_banditron} enables effective continuous learning for a dynamic iBMI system while maintaining low computational cost. Earlier work using Banditron\cite{shoeb_banditron} did not apply it to SNNs and was also limited to shallow networks that cannot provide sufficient task performance. In this work, we adapt the algorithm for SNNs and extend it to deep networks by training earlier layers using transfer learning. Using a spike based update rule keeps the computations in the update to be low as well.
 
\section{Self-Adaptive SNN decoder}
\label{sec:Self-AdaptiveSNNdecoder}
List of notations used in this paper
\begin{itemize}
    \item $N_{i}$: The number of neurons in $i^{th}$ layer
    \item $S_{i}$: Output of spiking neuron for $i^{th}$ layer
    \item $U_{i}$: Membrane potential of spiking neuron for $i^{th}$ layer
    \item $W_{i}$: Weight matrix for $i^{th}$ layer
    \item $N_{0}$: Number of Input Probes
    \item $T_W$: Bin window duration
    \item $St$: Stride size
    \item $s$: Sparsity
    \item $d$: Dropout rate
    \item $T_{GT_{open}}$: Sampling time period for ground truth labels in the open-loop experiment.
    \item $f_{GT_{open}}$: Ground truth label frequency ($=1/T_{GT_{open}}$)
    \item $T_{GT_{closed}}$: Sampling time period in the closed-loop experiment.
\end{itemize}
\subsection{Model Architecture}
\label{subsec:modelarchitecture}

The baseline DSNN model architecture is adapted from SNN\_streaming \cite{biyan2025combining} to comprise $k$ fully connected (FC) layers with neurons in each layer denotd by $N_i$, while inputs are denoted by $N_0$. Following \cite{biyan2025combining}, $k=3$ is selected and the final values of $N_i$ are optimized independently for open-loop and closed-loop experiments. 

\begin{figure}[!t]
\begin{framed}
    \centering
    \includegraphics[width=0.8\textwidth]{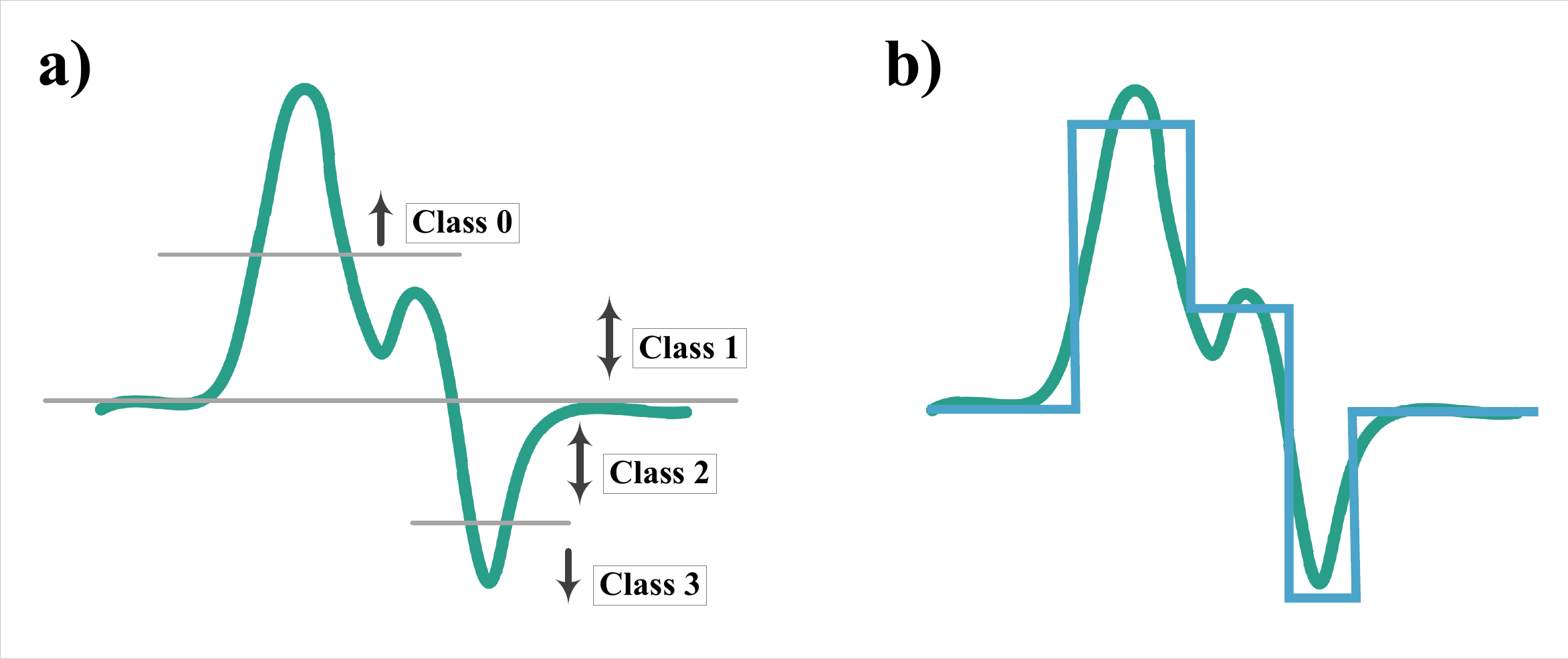}
    \caption{Transforming a regression task into a classification task. (a) Continuous target values are converted to discrete labels by grouping target values into bins. (b) Reconstruction of continuous output from discrete labels.}
    \label{fig:regression_classification}
\end{framed}
\end{figure}


All the SNN models used in this work utilize Leaky Integrate-and-Fire (LIF) neurons with subtractive reset. The corresponding equations for the LIF neuron are presented below:

\begin{align}
    U[t] &= \beta U[t-1] + WX[t] - S[t-1] \theta \nonumber \\
    \beta &= e^{-\Delta t/\tau} \nonumber \\
    S[t] &= \begin{cases}
        1 & \text{if } U[t] > U_{thr} \\
        0 & \text{otherwise }
    \end{cases} \nonumber \\
    \theta &= U_{thr} 
    \label{eq:lif_dynamic}
\end{align}

Where U[t], S[t], and X[t] are the membrane potential, output spike signal, and input signal at time t, while $\beta $ is the decay rate, W is the weight, $\theta$ is the reset mechanism, and $U_{thr}$ is the membrane potential threshold. To retain information from previous time steps, we implement subtractive reset in our models. The time step / bin window is set to the minimum value of $T_W=St=T_{GT_{open}}=4$ ms for the open-loop experiment and $T_W=St=T_{GT_{closed}}=10$ ms for the closed-loop experiment following \cite{taeckens2024spiking} (stride is set to same value for streaming data processing). For the first layer, X[t] is set to $1$ if there are any biological spikes in that time window and $0$ otherwise.

\begin{figure}[!t]
\begin{framed}
    \centering
    \includegraphics[width=0.98\textwidth]{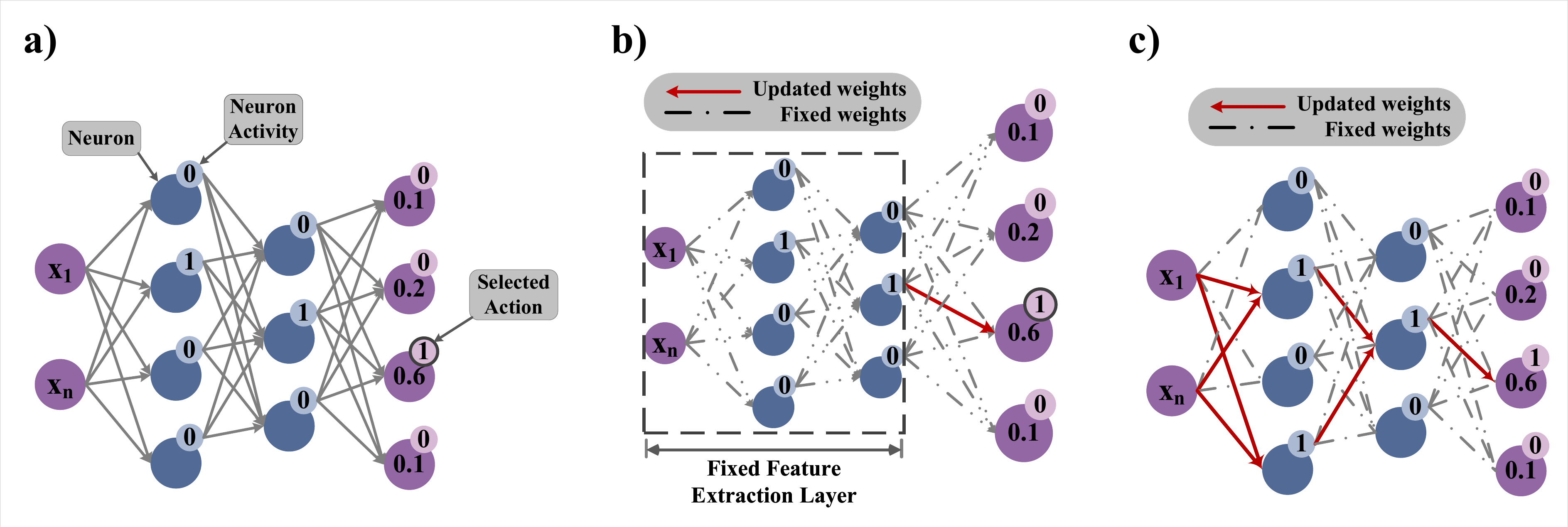}
    \caption{Illustration of the network architecture. Hidden neurons are represented in blue, while input and output neurons are shown in purple. The activity of neurons is indicated in the upper right circle of each neuron (a firing neuron is marked as 1, otherwise 0). (a) Forward pass of SNN. (b) Weight update pass for DSNN\_Banditron. (c) Weight update pass for DSNN\_AGREL.}
    \label{fig:forward_backward}
\end{framed}
\end{figure}

\subsection{Continuous Learning}
\label{subsec:continuouslearning}

 In most instances, the decoder models tend to remain fixed during inference in an iBMI system because of limited energy and computational resources \cite{9677062, shishavan2024closed}. However, many studies show that neural representations can drift over time \cite{sorrell2021brain}, the electrode may shift or be damaged during the real-life experiments \cite{rouanne2022unsupervised}. Therefore, if the decoder remains fixed, its performance may decline over time; therefore, the recalibration of decoders periodically is necessary. Nevertheless, many studies have found user independence to be a top priority for patients \cite{collinger2013functional}, it's not good to suspend the use of the decoder for retraining with the latest neural recording data from a user perspective. Thus, incorporating an energy-efficient online calibration is crucial for maintaining the decoder's optimal performance in real-life applications. We refer to the fixed DSNN model as baseline DSNN, and the model using the online training method as the ``DSNN\_X", where ``X" is the online learning method. 

 \begin{enumerate}

    \item \textit{Banditron}
    This efficient Banditron algorithm for multiclass classification in this work is adapted from \cite{kakade2008efficient,shoeb_banditron} as one of the  continuous learning methods. The first two layers of the DSNN are fixed as feature extraction layers based on transfer learning, while the weights between the third layer and output layer are adapted using the Banditron algorithm, as shown in \cref{fig:forward_backward}(b). We conducted our experiments in a bandit setting where the true label could not be exposed to the participant, only feedback was provided: $\mathbf{1}[S_{out}^{t}\neq y_t]$. In this equation, $S_{out}^{t}$ represents the predicted firing neuron, while $y_t$ is the actual label at time t. If the prediction does not match the label, the result of this equation is 1; otherwise, it is 0. Therefore, participants can only understand the correctness of the prediction, but cannot obtain the actual identity of the label. The prediction can be represented as below:
    
    \begin{align}
        S_{out}^{t} = \arg \max_{c\in[\mathcal{C}]}(U_{out}^{t})_{c} \nonumber \\
        U_{out}^{t} = \beta U_{out}^{t-1} + W^{t}S_{2}^t - S_{2}^{t-1} \theta 
    \label{eq:prediction}
    \end{align}

    where $(U_{out}^{t})_{c}$ represents the $c^{th}$ element of the membrane potential obtained from the spiking neuron in the output layer at time t, while $[\mathcal{C}]$ represents the set of $\mathcal{C}$ classes $[\mathcal{C}] = \{1,2,3,\cdots,\mathcal{C}\}$. Besides, $S_{2}^t$ denotes the spike output generated by the spike neurons in the second layer at time t, which subsequently serves as the input to the output layer, $\beta$ and $\theta$ are the decay rate and membrane threshold, respectively, as detailed in \cref{eq:lif_dynamic}. Moreover, the loss function is hinge loss because it is considered a convex loss function.
    
    \begin{align}
        \ell(W^t;S_{out}^{t}) = \max_{c\in[\mathcal{C}]/y_t}[1-(S_{out}^{t})_{y_t} + (S_{out}^{t})_{c}]_{+} * (dS_{out}^{t}/dU_{out}^{t})
    \label{eq:hinge}
    \end{align}

    The \cref{eq:hinge} illustrates the hinge loss of $W^{t}$ on $S_{out}^{t}$, the hinge function is $[\cdot]_{+}=\max\{\cdot,0\}$, which indicates the loss function is zero only if $(S_{out}^{t})_{y_t} - (S_{out}^{t})_{c} \geq 1$ for all $c\neq y_t$. However, when $S_{out}^{t}\neq y_t$ then $\ell(W^t;S_{out}^{t})\geq 1$, the hinge-loss is a convex upper bound on the zero-one loss function, $\ell(W^t;S_{out}^{t})\geq \mathbf{1}[S_{out}^{t}\neq y_t]$. For simplicity of implementation, we employ a straight-through estimator as a surrogate gradient defined as $dS/dU = 1$.

    Except for exploitation of $W^t$, Banditron also employs exploration with the probability of $1 - \epsilon$, where the parameter $\epsilon$ represents the exploration-exploitation trade-off. When conducting exploration, it uniformly predicts a random class from [$\mathcal{C}$] with the probability according to \cref{eq:explorationprob}, the exploration result denoted as $\tilde{y_t}$.

    \begin{align}
        \mathcal{P}(c) = (1-\epsilon)\mathbf{1}[c=S_{out}^{t}] + \frac{\epsilon}{\mathcal{C}}
    \label{eq:explorationprob}
    \end{align}

     When \(\tilde{y_t} \neq S_{out}^{t}\), the reception of positive feedback ($\tilde{y_t}={y_t}$) indicates the information regarding the identity of $y_t$, the update of the weight matrix $W^t$ can be conducted:

    \begin{align}
        W^{t+1} = W^t + \Delta W^t \nonumber \\
        \Delta W^t_{c,j} = S^{t}_{2,j}\left(\frac{\mathbf{1}[y_t = \tilde{y_t}]\mathbf{1}[\tilde{y_t}=c]}{\mathcal{P}(c)} - \mathbf{1}[S_{out}^{t}=c]\right)
    \label{eq:banditron_update}
    \end{align}

    where $\Delta W^t$ is the update matrix. And $c,j\in[\mathcal{C}],[N_{2}]$, where $[N_{2}]={ \{ 1,2,3,\cdots,N_{2} \} }$ denotes the set of neuron indices in the third layer.

    Note that the above description is valid for classification tasks only and hence earlier work\cite{shoeb_banditron} could not use it for velocity regression. In order to generalize this method to regression tasks, it is necessary to convert the original continuous velocity data into discrete class labels. This can be done by grouping the entire range of the target continuous variable into $B$ bins with different class labels, as illustrated in \cref{fig:regression_classification}(a) for $B=4$ and uniform quantization. These discrete labels are used to train the decoder. However, the continuous signal needs to be reconstructed during testing to evaluate the performance of the decoder and can be done using a zero order hold (\cref{fig:regression_classification} (b)). 

    \item \textit{Attention Gated Reinforcement Learning (AGREL)}  This algorithm was initially defined as a biologically plausible learning rule, and this type of learning rule is indeed implemented within the brain \cite{roelfsema2005attention, roelfsema2018control}. However, the original network only has a shallow hidden layer, \cite{pozzi2018biologically} extended this model to a deeper AGREL, and the algorithm described in this section is adapted from their work as shown in \cref{fig:forward_backward} (c). In contrast to Banditron, which only updates the final layer, AGREL performs updates across all layers of the model. Synaptic plasticity is determined by two key factors. The first is the reward prediction error $\delta$, which is computed globally after the action is evident. It is positive when the action selected by the network results in an increased reward; otherwise, it is negative. The second factor is the attentional feedback signal, which is transmitted from the output layer back to the previous layers. The learning rule can be represented as follows on the $i^{th}$ layer:

    \begin{align}
        \Delta W_{i}[t] = \alpha*S_{i-1}[t] \cdot fb_{i}[t] \nonumber \\
        fb_{i}[t] = e_{i}[t]*S_{i}[t]*(dS_{i}[t]/dU_{i}[t]) \nonumber \\
        e_{i}[t] = fb_{i+1}[t] \cdot W_{i+1}[t]
    \label{eq:AGREL_update_layeri}
    \end{align}

    where $\alpha$ represents the learning rate, $S_{i}[t]$ denotes the spike output generated by the spike neurons in the $i^{th}$ layer at time t. To address the dead neuron problem, we use a straight-through estimator as a surrogate gradient to estimate the gradient of the spiking neuron: $dS/dU = 1$. This has the advantage of simple implementation and also provided similar performance to other complex surrogate functions such as arctan. The variable $fb_{i}$ refers to the feedback signal, while $e_{i}$ indicates the error signal received by layer $i$. However, the update mechanism for the last layer is distinct, as the error signal is substituted with the reward prediction error:

    \begin{align}
        \Delta W_{out}[t] = \alpha*S_{i-1}[t] \cdot fb_{out}[t] \nonumber \\
        fb_{out}[t] = \delta[t]*z_{out}[t]*(dS_{i}[t]/dU_{i}[t])
    \label{eq:AGREL_update_layerlast}
    \end{align}

    where $z_{out}$ indicates the neuron activity of the output layer, the output assigned as the `winning' unit receives an activity value of 1, while all other units are assigned a value of 0; the prediction can also be written as \cref{eq:prediction}.
    
    If the network selects the correct output unit, the reward signal r is assigned as 1, the reward prediction error $\delta$ can be computed as $\delta = r-z_{c}$, where $z_{c}$ is the activity of the winning unit. If misclassified, no reward is assigned (r = 0), while $\delta = -1$. The \cref{eq:AGREL_update_layeri} and \cref{eq:AGREL_update_layerlast} indicate that only the synaptic connections between the winning unit are updated, because the other units have activity 0 after the competition (as shown in \cref{fig:forward_backward} (c)). Thus, the feedback signal gates the plasticity of connections from the previous layer to the next layer, thereby assigning credit to hidden units that are responsible for the choice of action.
    
\end{enumerate}

\section{Methodology}
\label{sec:Methodology}

We conducted open-loop and closed-loop experiments independently and will describe each type separately in this section. Open-loop experiments contribute to our understanding of how the brain encodes movement and play a vital role in the design of decoders \cite{sorrell2021brain}, while closed-loop experiments with perturbations better simulate real-life situations and facilitate the evaluation of the decoder's robustness.

\subsection{Open-loop Dataset}
\label{subsec:openloopdataset}
The primate reaching dataset \cite{makin2018superior} selected for this paper was recorded from the brains of two macaque monkeys, Indy and Loco. The monkey performed the reach task by moving a cursor towards a randomly generated target, while recordings were made from a microelectrode array (MEA) for $N_{0} = $ 96 channels or two MEAs with $N_{0} = $ 192 channels. The reach task is organized on an 8x8 grid and is performed by one of the monkey's arms (shown in \cref{fig:openclosedloop} (a)). When the cursor reaches the target and remains there for a certain duration, the target location changes. The moving velocity of the cursor is referred to as the label in our experiment, and the spike data are unsorted, and both were sampled at $f_{GT_{open}}$ = 250 Hz. In this paper, we select six recordings from Indy that were recorded by M1 arrays on \textbf{adjacent dates} (“indy\_20161005\_06”, "indy\_20161006\_02", "indy\_20161007\_02", "indy\_20161011\_03", "indy\_20161013\_03", "indy\_20161014\_04") as the open-loop dataset. This is different from the selection in Neurobench\cite{yik2025neurobench} since in that case the objective was to span the entire recording range of several months, and retraining using daily data was allowed. In contrast, we do not allow usage of data beyond the first day and hence use data from nearby days to maintain some level of correlation across days. The first 80\% of the recording from the earliest file ("indy\_20161005\_06") is used for training data, while the remaining recordings are used for testing. In the data pre-processing stage, we used a 4 ms bin window to count the number of spikes in that period \cite{yik2025neurobench}.

\subsection{Closed-loop Data and Experiments}
\label{subsec:closedloopexperiments}

Compared with an open-loop system, a closed-loop system is closer to real life, as it allows patients to provide feedback to the system and make attempts to rectify any errors in the output. However, experiments involving human participants require a significant investment of both financial resources and time. An alternative method is to employ an online prosthetic simulator (OPS) \cite{cunningham2011closed} to simulate the neural responses of a real human in the closed-loop system. This simulator assigns a distinct preferred direction to each neuron and uses the cosine tuning model \cite{georgopoulos1986neuronal} to adjust their firing rates accordingly. Hence, neurons are more likely to fire when the intended motion direction is close to their preferred direction: 

\begin{align}
    \lambda_{t} = (\lambda_{max} - \lambda_{min})c_{k}\cdot x_{t}+\lambda_{min}
    \label{eq:ops}
\end{align}

Where $\lambda_{t}$ represents the firing rate at time t, while $\lambda_{max}$ and $\lambda_{min}$ are the maximum and minimum firing rate, respectively. We defined that $\lambda_{min}$ can be sampled from 0 to 5 spikes per second, while $\lambda_{max}$ can be sampled from 40 to 100 spikes per second, as these values correspond to the firing rates observed in real neural data \cite{taeckens2024spiking}. Additionally, $c_{k}$ represents the preferred direction of the $k^{th}$ neuron, and $x_{t}$ denotes the intended acceleration vector. The spiking probability, represented as $p=\lambda_{t}*T_{GT_{closed}}$, determines the number of spikes generated within each sampling interval. To create a more realistic simulation, random noise with a normal distribution (N(0,0.3)) is incorporated into the OPS brain model.

The closed-loop experiment employs the center-out task as demonstrated by \cite{cunningham2011closed}, while the animal participants are replaced by a brain model (OPS). In this virtual environment, all actions are performed on a screen. The starting point is located at the center of the screen, and a target is randomly generated around it at a fixed distance. As illustration in \cref{fig:openclosedloop} (b), the desired acceleration is sent to the OPS to generate the brain signal, indicating that the virtual participant can see the target's position and attempt to move towards it. Then, the brain signals are sent to a decoder, and the environment receives the predicted velocity from the decoder to update its status. Additionally, the environment provides participants with a feedback signal, referred to as a reward signal, which indicates the participant can observe changes and evaluate the correctness of those changes within the environment. The reward signal can be employed to calibrate decoders or to evaluate the performance of the system.

A reach task is determined as successful if it meets two conditions simultaneously (same as \cite{cunningham2011closed}): first, the time to target (the time taken from the start point to the target) is less than the maximum allowable duration (3 s); second, the cursor needs to stay within an acceptance window for at least 0.5 s. The acceptance window is defined as a circle with a radius of 4 unit distances around the target.

\subsection{Metrics}
\label{subsec:metrics}

To comprehensively evaluate the performance of models in terms of cost vs. accuracy, several metrics are used following Neurobench \cite{yik2025neurobench}: synaptic operations, memory footprint, and accuracy. For synaptic operations, two key metrics need to be considered: multiply-and-accumulate (MAC)/accumulations (ACs) and memory access. The MAC operation is a widely utilized metric in ANN, which is employed to quantify the total number of operations involving multiplication followed by accumulation, serving as an important indicator of computational efficiency and performance. However, when neuron activation is constrained to binary values (such as SNN), multiplication operations are excluded, and the corresponding metric is referred to as ACs. Additionally, memory access is utilized to count read and write operations within memory. This is considered a metric because the energy consumed by memory access often dominates that used for computations \cite{horowitz20141}. Moreover, since we consider adaptive methods which require writes to memory during parameter updates, this metric becomes an important comparison point between adaptive algorithms. Hence, we consider memory accesses separately for forward and backward passes, a consideration not present in earlier benchmarking work \cite{yik2025neurobench}. The memory footprint is used to determine the memory size required for models, where parameters are stored as 32-bit floating-point numbers. Furthermore, different criteria are employed to evaluate accuracy for open-loop and closed-loop systems. The coefficient of determination, denoted as $R^2$, is a widely utilized metric in regression tasks \cite{yik2025neurobench}, and it serves as the accuracy metric in open-loop experiments:

\begin{equation}
    R^{2} = 1-\frac{ {\textstyle \sum_{i=1}^{n}\left ( y_{i}-\hat{y_{i} }  \right )^{2}  } }{{\textstyle \sum_{i=1}^{n}\left ( y_{i}-\bar{y} \right )^{2}  }} 
    \label{eq:r2}
\end{equation}

In this equation, \(\hat{y}_i\) represents the predicted values, \(y_i\) refers to the actual labels, and \(\bar{y}\) indicates the mean value of the labels. In our experiments, the output is calculated independently in the x and y directions. Therefore, the final value of $R^{2}$ is determined as the average of $R_x^{2}$ and $R_y^{2}$.

In the closed-loop experiments, the performance of the system is assessed using a metric known as time-to-target, referring to the time required to reach the target \cite{cunningham2011closed}.

\subsection{Training Details}
\label{subsec:training_details}

The models are trained independently in open-loop and closed-loop experiments, each consisting of two training phases: pre-training and post-training. Pre-training aims to establish the foundational capabilities of the model at the start of the experiments, while post-training is performed during the model's inference stage, with continuous learning as the primary methodology. For the testing part, all models are tested with streaming data (a continuous data flow without mini-batch), as this is more representative of natural data flow \cite{wairagkar2025instantaneous} \cite{littlejohn2025streaming}. This section provides an introduction to model pretraining, while the continuous learning algorithm is discussed in \Cref{subsec:continuouslearning}.

In open-loop mode, models are pre-trained for 50 epochs using the SNNTorch framework, with AdamW as the optimizer and a learning rate of 0.01. Cross-entropy loss is employed as the loss function, and the batch size is set to 512. The surrogate gradient utilized in this stage is the arctan function \cite{fang2021incorporating}.

For closed-loop, the pre-training has two stages, since two-stage training can enhance the robustness of the online learning models \cite{willsey2022real}. During the first stage,  the training process is the same as in open-loop, but the learning rate is set to 0.005, with 5 epochs for the baseline DSNN and 10 epochs for the one-layer Banditron. In the second stage, the brain data generated by OPS is fed into the pre-trained DSNN model, which provides an intended direction of movement that moves the cursor, resulting in a new set of brain data generated by OPS. These trials are utilized to further update the model weights in an online fashion, while the learning rate in this stage is changed from 5e-8 to 5e-10. All closed-loop experiments are performed after the two-stage training.

\section{Results}
\label{sec:Results}

To comprehensively examine the capability of different algorithms, we performed multiple experiments and evaluated models using the metrics mentioned in \cref{subsec:metrics}. The ``DSNN" in this section refers to the baseline DSNN model without implementing continuous learning during the inference stage, whereas ``DSNN\_Banditron/AGREL" indicates Banditron and AGREL as continuous learning algorithms utilized during the inference process. ``CLSNN" refers to the previous work in \cite{taeckens2024spiking} that uses SNN with continual learning. Additionally, ``Banditron" denotes the original Banditron algorithm \cite{kakade2008efficient}, which is a simple network with only one layer.

\subsection{Open-loop Results}
\label{subsec:open-loop_results}

\begin{figure}[!t]
\begin{framed}
    \centering
    \begin{subfigure}[b]{0.6\textwidth}
        \centering
         \includegraphics[width=\textwidth]{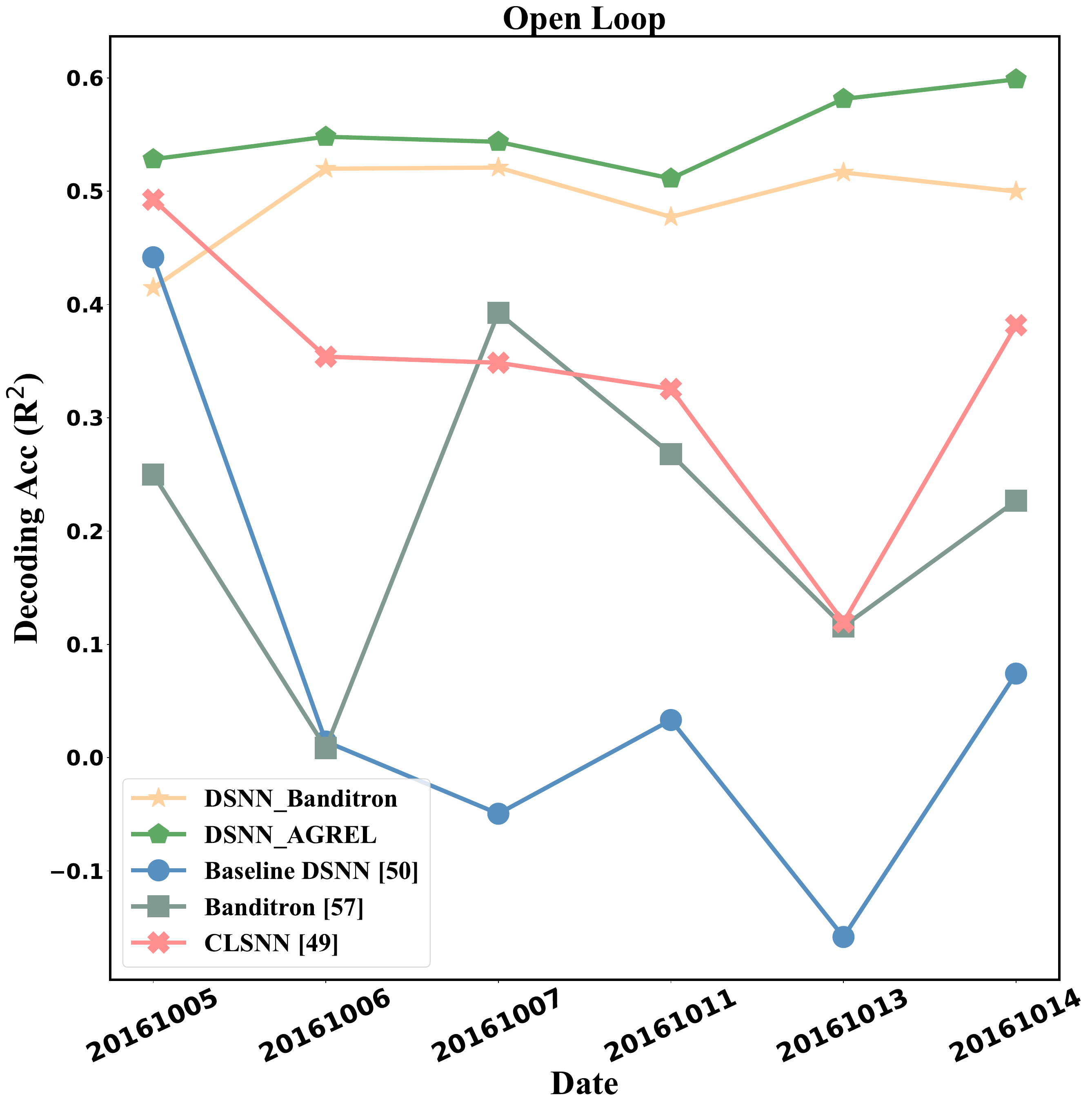}
        \caption{}
    \end{subfigure}
    \caption{The comparison of continuous learning performance in Open-loop experiments across several days. The accuracy of DSNN\_AGREL and DSNN\_Banditron is the most stable across days and retains high accuracy across all days.}
    \label{fig:acc_tra}
\end{framed}
\end{figure}
In the open-loop experiments, the optimization of the network architecture was performed using the grid search method on the dataset file “indy\_20161005\_06.” Consequently, the resulting network architecture comprises two hidden layers, each containing 30 neurons (i.e. $N_1=N_2=30$). Additionally, dropout layers with a dropout rate of d = 0.1 are incorporated following each hidden layer to enhance model robustness. Furthermore, the input layer has been modified to accommodate the current dataset of 96 channels, i.e. $N_0=96$. However, the output layer now consists of $N_3=8$ neurons (4 classes each for x and y direction), as the original regression task has been transformed into a classification task.

The dataset used in the open-loop experiment is detailed in \cref{subsec:openloopdataset}, while the pre-training method is explained in \cref{subsec:training_details}. The open-loop experiments are performed during the model inference phase, employing streaming data for analysis. Four different algorithms were employed in the open-loop experiment, and the result is illustrated in \cref{fig:acc_tra}.  It is evident that the networks without online calibration perform poorly compared to those with online learning; in fact, the decoding accuracy nearly drops to zero by the second day after pre-training. Over the following days, it remains close to zero, with the highest accuracy only approximately 0.1. This phenomenon demonstrates that the accuracy of decoding is not consistent over time. Therefore, recalibration or online learning is necessary. The original Banditron outperforms baseline DSNN, but its accuracy is unstable, the highest accuracy can reach about 0.4, while the lowest is around 0. In contrast, DSNN\_Banditron and DSNN\_AGREL show stable performance over several days, achieving the highest accuracy during that period. Additionally, most of the time both DSNN\_AGREL and DSNN\_Banditron have accuracies between 0.5 and 0.6, while DSNN\_AGREL shows a slightly higher accuracy albeit at a much higher computation cost as shown later. 

\subsection{Closed-loop Results with Perturbations}
\label{subsec:closed-loop_results}

\begin{figure}[!t]
\begin{framed}
    \begin{subfigure}[b]{0.48\textwidth}
        \centering
         \includegraphics[width=\textwidth]{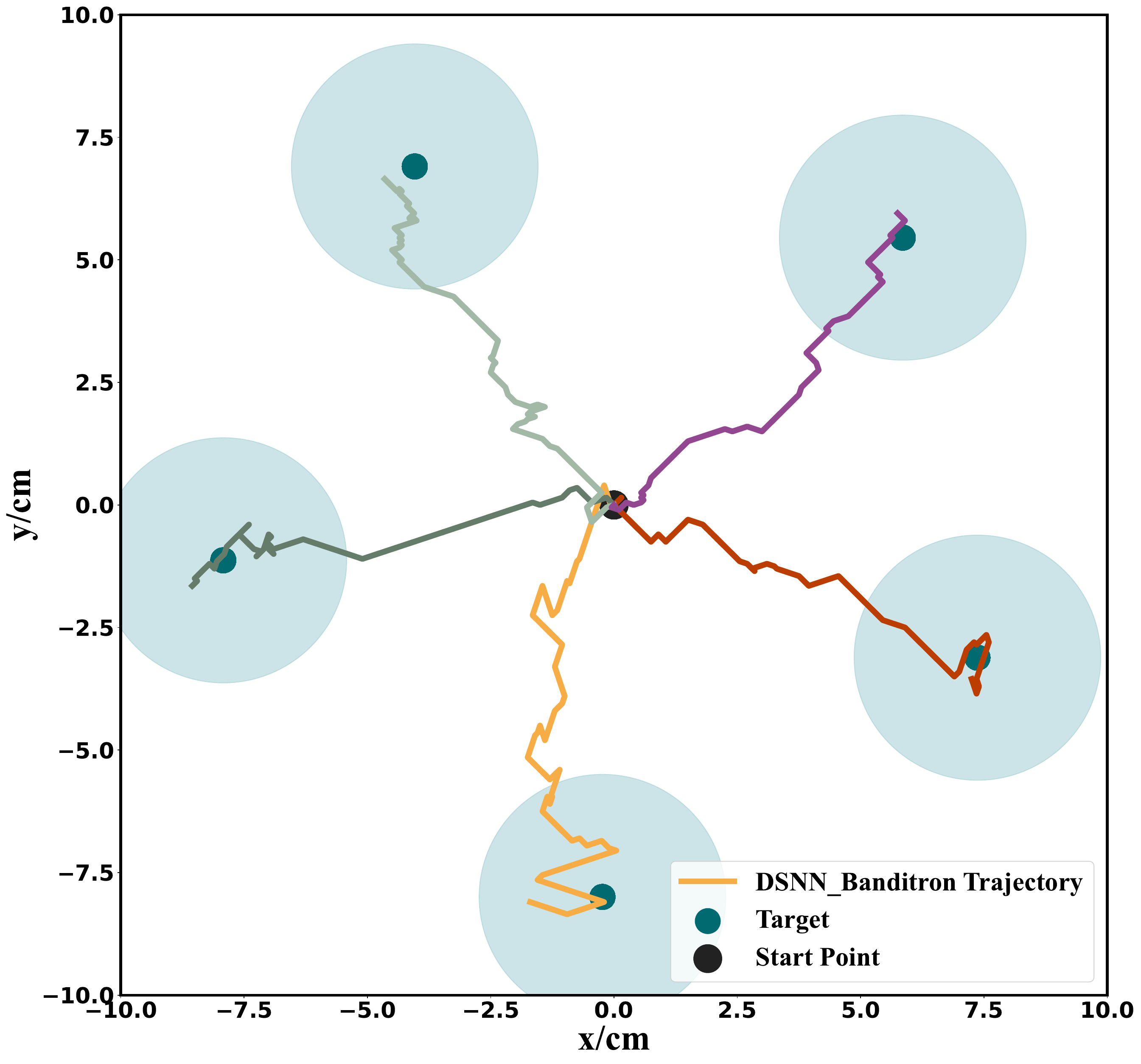}
        \caption{}
    \end{subfigure}
    \hfill
    \begin{subfigure}[b]{0.48\textwidth}
        \centering
         \includegraphics[width=\textwidth]{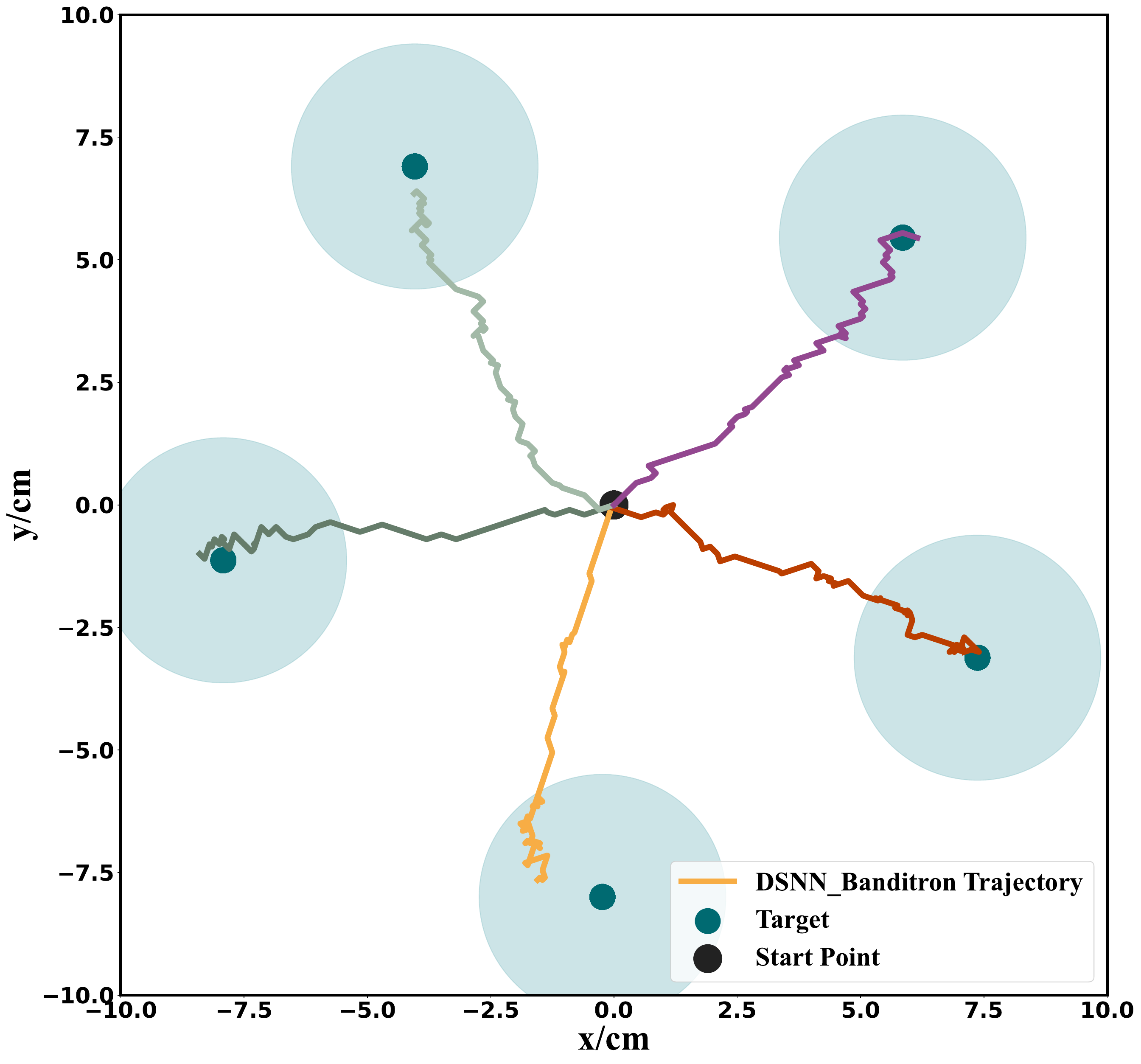}
        \caption{}
    \end{subfigure}
    \caption{(a) The trajectory of the closed-loop experiment right after loss of neuron with a perturbation ratio of 0.6. (b)  The trajectory of the closed-loop experiment with loss of neuron after continuous learning for 50 trials has reduced time to target.}
    \label{fig:closed_traj_ln}
\end{framed}
\end{figure}

\begin{figure}[!t]
\begin{framed}
    \centering
    \begin{subfigure}[b]{0.32\textwidth}
        \centering
         \includegraphics[width=\textwidth]{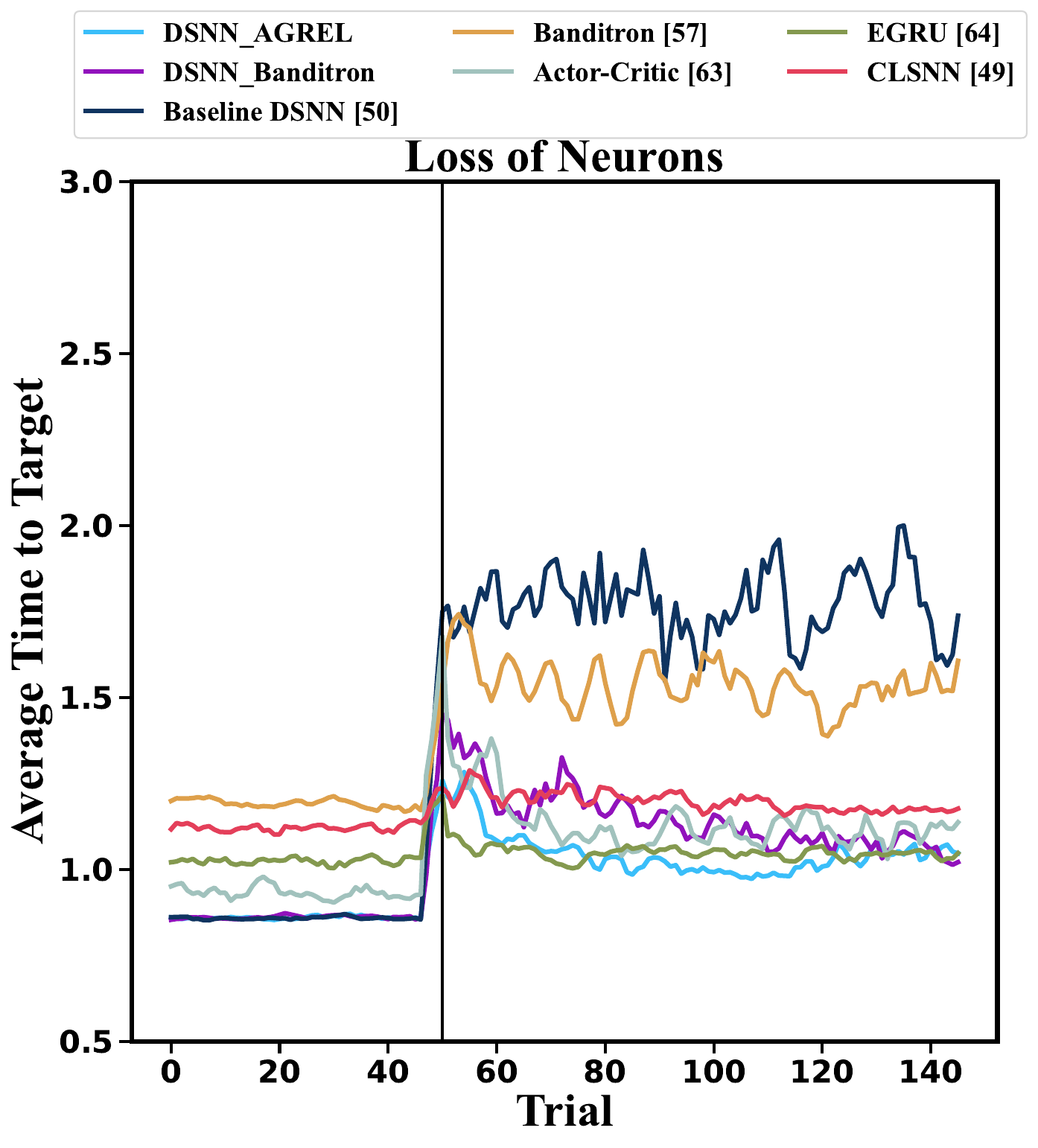}
        \caption{}
    \end{subfigure}
    \hfill
    \begin{subfigure}[b]{0.32\textwidth}
        \centering
         \includegraphics[width=\textwidth]{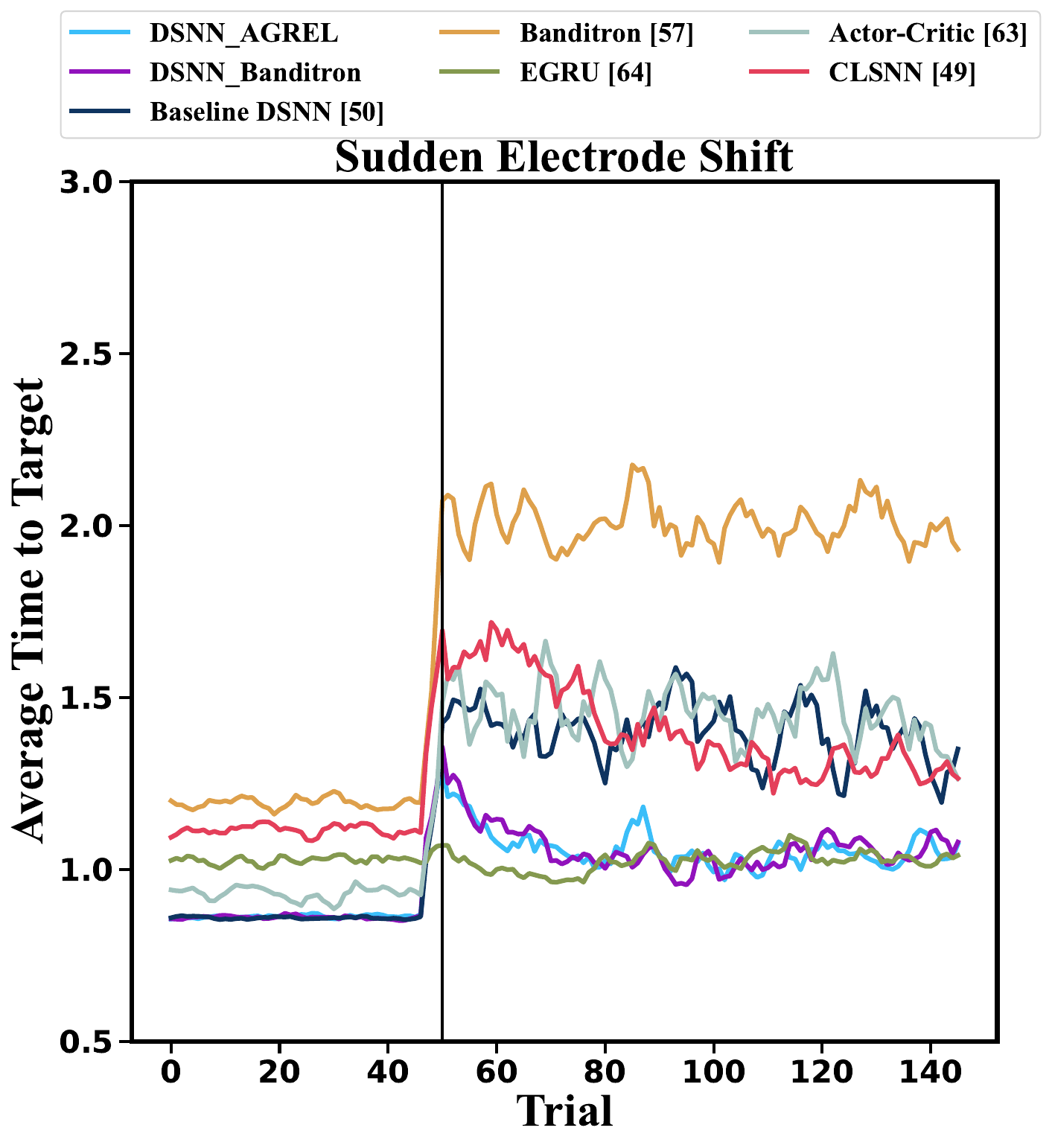}
        \caption{}
    \end{subfigure}
    \hfill
    \begin{subfigure}[b]{0.32\textwidth}
        \centering
         \includegraphics[width=\textwidth]{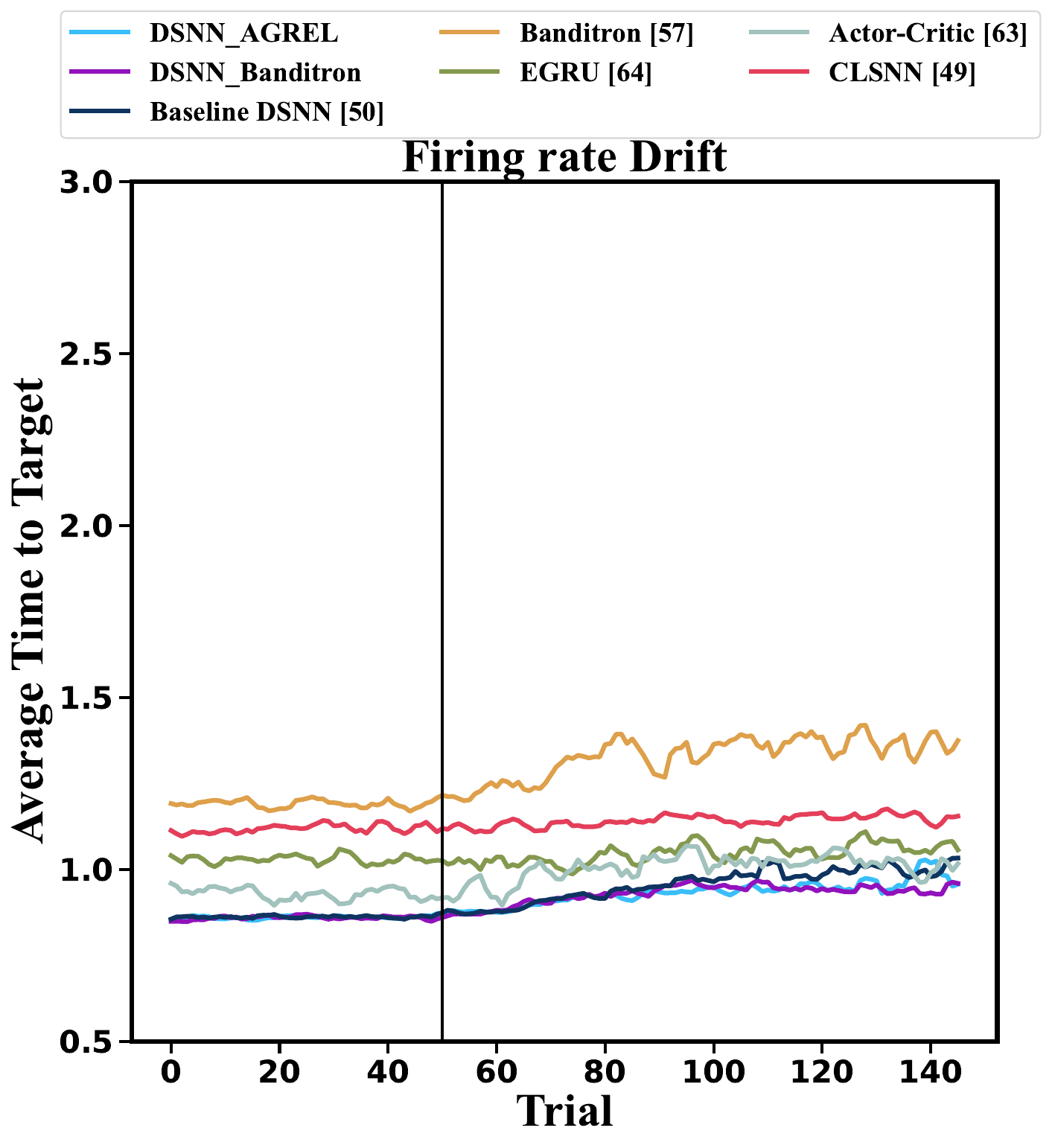}
        \caption{}
    \end{subfigure}
    \caption{The average time to target for perturbation experiments: (a) Loss of neurons. (b) Sudden electrode shift. (c) Firing rate drift. DSNN\_AGREL, DSNN\_Banditron and EGRU show better performance among the adaptive methods while Banditron is the worst in terms of adapting back to baseline after perturbation.}
    \label{fig:perturbations}
\end{framed}
\end{figure}
In closed-loop experiments, there are $N_0=46$ input neurons, to align with the experiments in \cite{taeckens2024spiking}. Additionally, the architecture consists of two hidden layers with 65 and 40 neurons, respectively (i.e. $N_1=65$ and $N_2=40$). The output layer has also been adjusted to consist of $N_3=8$ neurons, with 4 neurons for each of the x and y directions. Similar to the open-loop configuration, two dropout layers are employed after each hidden layer, utilizing a dropout rate of d = 0.3.

The closed-loop experiments are conducted after the 2-stage training, as detailed in \cref{subsec:training_details}. Apart from the normal closed-loop experiments, some perturbations are introduced in this section to simulate various real-life situations following the methods in \cite{taeckens2024spiking} for one to one comparisons. The performance of the experiments is assessed by calculating the average time-to-target across 10 simulations, with neurons being selected randomly in each simulation. As shown in \cref{fig:perturbations}, the first 50 trials in every experiment are conducted in the normal closed-loop experiments, while perturbations are performed afterwards.

The first experiment is the loss of neurons. After the first 50 trials, 30 neurons are randomly selected from the total of 46 neurons and removed. This experiment simulates the loss of information in the real-life iBMI system when electrodes become disconnected from neurons due to damaged or shifted \cite{rouanne2022unsupervised}. The trajectory of the DSNN\_Banditron in \cref{fig:closed_traj_ln}(a) shows a curving path after this perturbation, indicating that it requires a long time to reach the target. The different colors indicate the trajectories from the start point to various targets. The \cref{fig:closed_traj_ln} (b) illustrates the trajectory after adaptation for 50 trials in a loss of neurons experiment. It is apparent that the trajectories of DSNN\_Banditron are smooth after 50 trials demonstrating the effect of learning qualitatively.

For a more quantitative analysis, \cref{fig:perturbations} (a) depicts the average time to target has a sudden increase at the $50^{th}$ trial. The time-to-target for DSNN\_AGREL is approximately 1 s after around 30 trials, while the learning speed of DSNN\_Banditron is relatively slower compared to DSNN\_AGREL. However, it eventually converges to the same level as DSNN\_AGREL at about 60 trials.  In contrast, the baseline DSNN maintains a high time-to-target without any noticeable decrease,  while the Banditron shows a slight decreasing trend from about 1.7 s to 1.5 s, although this trend is not very obvious.

The second experiment simulates a sudden shift in position of electrodes. The preferred direction of 30 randomly selected neurons is reassigned. In the real-life iBMI system, this situation primarily occurs when the electrode shifts to another neuron and obtains information from the new target neuron \cite{rouanne2022unsupervised}. \cref{fig:perturbations} (b) indicates that there is also a notable change in the performance of the models around the $50^{th}$ trial. The performance for DSNN\_AGREL and DSNN\_Banditron is similar, both returning to 1.0 seconds in approximately 30 trials. However, the average time-to-target for both the Banditron and the baseline DSNN demonstrates no significant reduction.

The third experiment is the firing rate drift. In this experiment, the maximum firing rate for 30 randomly selected neurons was reassigned from a range of 40-100 spikes per second to a new range of 0-30 spikes per second. Many study shows that even in stable behaviors, the neural representations change \cite{sorrell2021brain} or the firing rates of neurons drift over time \cite{ganguly2009emergence}. The performance of all models does not change significantly during this experiment, exhibiting only a slight increase after about $60^{th}$, as demonstrated in \cref{fig:perturbations} (c). Among the models, Banditron demonstrates the most significant variability, indicating that it is more susceptible to variations in environmental conditions. DSNN\_AGREL and DSNN\_Banditron also show a slight increase but recover after several trials. 

\begin{figure}[!t]
\begin{framed}
    \centering
    \includegraphics[width=0.98\textwidth]{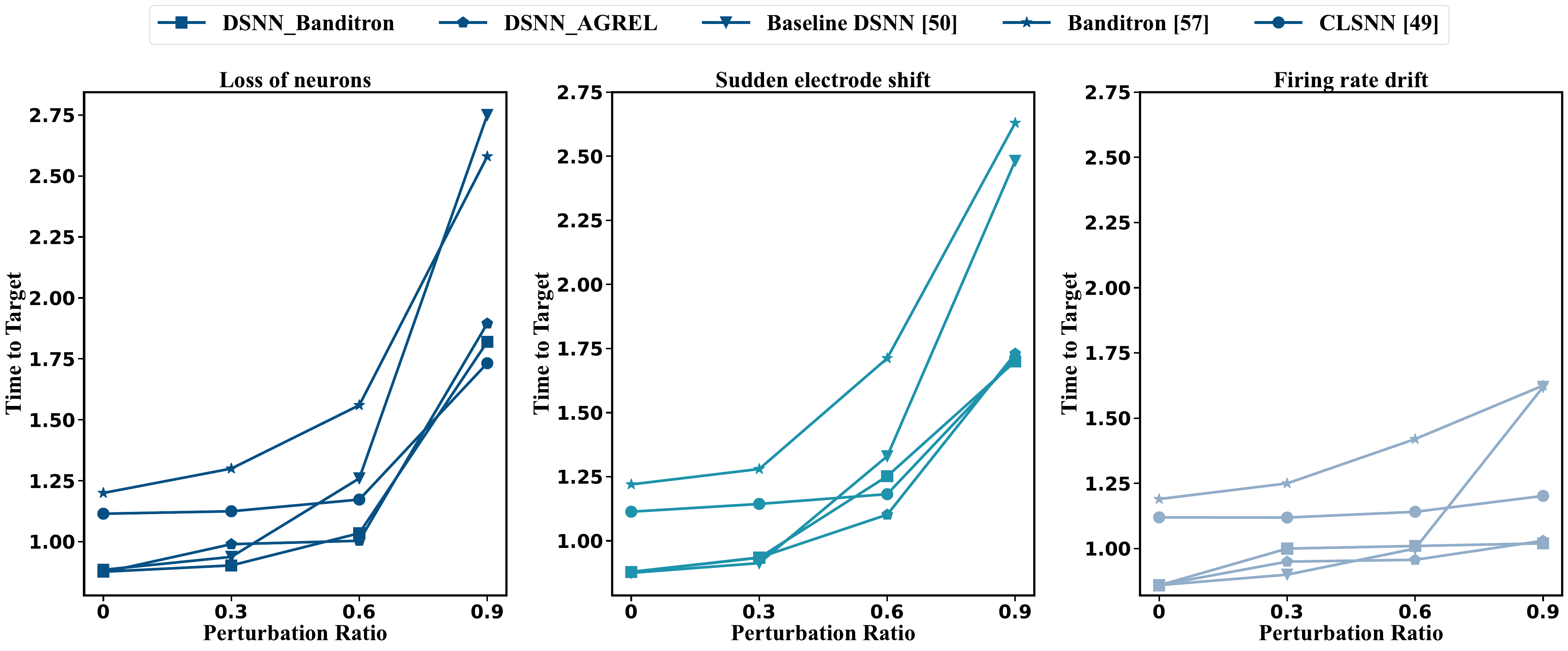}
    \caption{The time to target for three types of perturbations, with the ratios of perturbed neurons ranging from 0\% to 90\%. The continuous learning models retain performance across a wide range of perturbation ratios, with the original Banditron performing worst.}
    \label{fig:perturbationratio}
\end{framed}
\end{figure}

 Furthermore, \cref{fig:perturbationratio} presents an analysis of the models' performance in relation to the varying number of neurons affected by perturbation. The ratio of the number of perturbed neurons to total neurons is referred to as perturbation ratio. When the perturbation ratio is low, the performance of all models across all types of perturbations is not significant. However, it is evident that when the perturbation ratio exceeds 60\%, the self-adaptive strategies demonstrate a substantial advantage in performance restoration compared to the baseline DSNN. Although the performance of DSNN\_Banditron is comparable to that of DSNN\_AGREL, DSNN\_AGREL demonstrates greater stability in performance when subjected to a higher ratio of neuron perturbation.

\subsection{Resource Requirements vs. Performance}
\label{subsec:costvsacc}

The implanted decoder in real-world applications requires an extremely energy-efficient and area-efficient \cite{shaeri202433, cindy_tbcas}, as well as maintaining a high level of accuracy; therefore, the analysis of the resource requirements of the models and algorithms is necessary. The metrics used in this section are described in \cref{subsec:metrics} and the Neurobench software \cite{yik2025neurobench} is used to obtain these metrics. The operations are calculated separately in the forward path and the backward path due to our work also accounting for parameter updates, and the results are demonstrated in \cref{tab:computes}. The "Average Time" in \cref{tab:computes} represents the average time-to-target from the $50^{th}$ trial to the $100^{th}$ trial from the start of the three types of perturbations across ten simulations. This reflects the performance of models after the post-training. Furthermore, the analyses regarding the operations and memory requirements for DSNN\_Banditron, DSNN\_AGREL, and Banditron are comprehensively detailed in \cref{sec:weightupdate}, while information listed for the other models has been derived from the source code, and a straightforward analysis is also presented in \cref{sec:weightupdate}. 

DSNN\_AGREL achieves the best average time to target among models, while DSNN\_Banditron is ranked second with only a 0.01s difference from DSNN\_AGREL. The performance of baseline DSNN illustrates the consequences of maintaining a fixed network throughout all experiments, and its poor performance highlights the necessity for continuous learning in complex environments. For the forward pass, the computation depends on the model structure. The continuous learning SNN from \cite{taeckens2024spiking} has a similar model structure to the model in this work, but it has a slightly smaller memory footprint and forward computation because it only utilizes $N_3=2$ output neurons for a regression task whereas we employed $N_3=8$ neurons due to converting the regression to classification tasks. The actor-critic reinforcement learning model from \cite{7252} consists of two models: the actor model and the critic model. Since the critic model is not used during inference, the forward computations only involve the actor model. EGRU \cite{evnn2023} is a highly efficient model, known for its sparse inference and superior performance. Its forward computations are fewer than most of the models listed in the table. All of these models have inferior performance compared to the proposed DSNN\_X models.

\begin{table}[!t]
\centering
\caption{Resource requirement of models/methods}
\scriptsize
\begin{tabular}{@{}llllllllllll}
\br
\centre{1}{\textbf{Models}}&\centre{1}{\textbf{Average}}&&\centre{1}{\textbf{Average}}&&&\centre{1}{\textbf{Forward}}&&&\centre{2}{\textbf{Backward}}& \centre{1}{\textbf{Memory}} \\

\centre{1}{\textbf{/Methods}} & \centre{1}{\textbf{Time}} && \centre{1}{\textbf{Time}} & & & \centre{1}{\textbf{Computes}}&  & & \centre{2}{\textbf{Computes}}  & \centre{1}{\textbf{Footprint $($ kB $)$}} \\

&\centre{1}{\textbf{(s)}} &&\centre{1}{\textbf{Perturb}}&&\crule{3} & &\crule{2} \\

&&&\centre{1}{\textbf{(s)}}&&\centre{1}{\textbf{MACs}} &\centre{1}{\textbf{ACs}} &\centre{1}{\textbf{Memory}}& &\centre{1}{\textbf{MACs}} &\centre{1}{\textbf{Memory}}\\

&&&&&&& \centre{1}{\textbf{Access}} &&& \centre{1}{\textbf{Access}} \\

\mr

\centre{1}{DSNN\_Banditron} & \centre{1}{{0.87}} && \centre{1}{{1.02}} & & \centre{1}{{113}} & \centre{1}{{2590.0}} & \centre{1}{{2590.0}} & & \centre{1}{{32.0}}& \centre{1}{{32.0}}& \centre{1}{{24.54}}\\

\mr

\centre{1}{DSNN\_AGREL} & \centre{1}{{0.87}} && \centre{1}{{1.01}} & & \centre{1}{{113}} & \centre{1}{{2590.0}} & \centre{1}{{2590.0}} & & \centre{1}{317.04}& \centre{1}{{335.84}}& \centre{1}{{24.54}}\\

\mr

\centre{1}{baseline DSNN\cite{biyan2025combining}} & \centre{1}{{0.87}} && \centre{1}{{1.38}} & & \centre{1}{{113}} & \centre{1}{{2590.0}} & \centre{1}{{2590.0}} & & \centre{1}{{0}}& \centre{1}{{0}}& \centre{1}{{24.54}}\\

\mr

\centre{1}{Banditron \cite{kakade2008efficient}} & \centre{1}{{1.21}} && \centre{1}{{1.63}} & & \centre{1}{{147.2}} & \centre{1}{{0}} & \centre{1}{{147.2}} & & \centre{1}{{36.8}}& \centre{1}{{36.8}}& \centre{1}{{1.4375}}\\
\mr

\centre{1}{Continuous learning} & \centre{1}{{1.12}} && \centre{1}{{1.22}} & & \centre{1}{{107}} & \centre{1}{{2534.0}} & \centre{1}{{2534.0}} & & \centre{1}{{5020.0}}& \centre{1}{{2510.0}}& \centre{1}{{24.06}}\\
\centre{1}{SNN \cite{taeckens2024spiking}} & \centre{1}{{}} && \centre{1}{{}} & & \centre{1}{{}} & \centre{1}{{}} & \centre{1}{{}} & & \centre{1}{{}}& \centre{1}{{}}& \centre{1}{{}}\\

\mr

\centre{1}{Actor-Critic RL \cite{7252}} & \centre{1}{{0.94}} && \centre{1}{{1.19}} & & \centre{1}{{96}} & \centre{1}{{624.8}} & \centre{1}{{624.8}} & & \centre{1}{{154598}}& \centre{1}{{77299}}& \centre{1}{{229.8}}\\

\mr

\centre{1}{EGRU \cite{evnn2023}} & \centre{1}{{1.03}} && \centre{1}{{1.05}} & & \centre{1}{{288}} & \centre{1}{{0}} & \centre{1}{{322}} & & \centre{1}{{1252}}& \centre{1}{{626}}& \centre{1}{{1.3}}\\

\br
\end{tabular}
\label{tab:computes}
\end{table}

It is evident that the most energy-efficient method for the backward pass is the learning methodology associated with Banditron. However, the performance of the single-layer Banditron is constrained by its limited generalization capability, so the incorporation of an additional feature extraction layer is essential for stable performance, as done in our proposed DSNN\_Banditron. The computational requirements of DSNN\_AGREL are significantly higher than those of DSNN\_Banditron. This discrepancy arises from the fact that DSNN\_AGREL updates weights across every layer, while DSNN\_Banditron performs updates in only a single layer. The continuous learning SNN requires more backward computations than DSNN\_AGREL, which is due to lack of sparsity  in its update mechanism during the backward phase. The actor-critic model requires large computations in the backward path, since the critic model also needs to be learned and updated along with the actor model. The EGRU model exhibits a greater computational demand in the backward path compared to the forward path, primarily due to the increased computational requirements associated with the EGRU cell. In summary, DSNN\_Banditron achieves the most energy-efficient method of continuous learning. The analysis details of computations/memory requirements of the continuous learning SNN from \cite{taeckens2024spiking}, the actor-critic RL algorithm from \cite{7252}, and the EGRU from \cite{evnn2023} are briefly explained in \cref{sec:weightupdate}.

\section{Discussion}
\label{sec:Discussion}
We presented a low-complexity DSNN decoder for iBMI that can adapt to changing neural data statistics by using a simple RL algorithm known as Banditron. Banditron requires change of weights in only one layer. Prior work using shallow one-layer decoders are however unable to perform well in the face of large perturbations. In contrast, we show deep SNNs can learn some invariant features which can transfer well to other scenarios when the neural statistics are perturbed. We discuss some further critical aspects about the work below.

\subsection{Comparison with the Linear Decoder}
\label{sec:lineardecoder}

\begin{figure}[!t]
\begin{framed}
    \centering
    \includegraphics[width=0.6\textwidth]{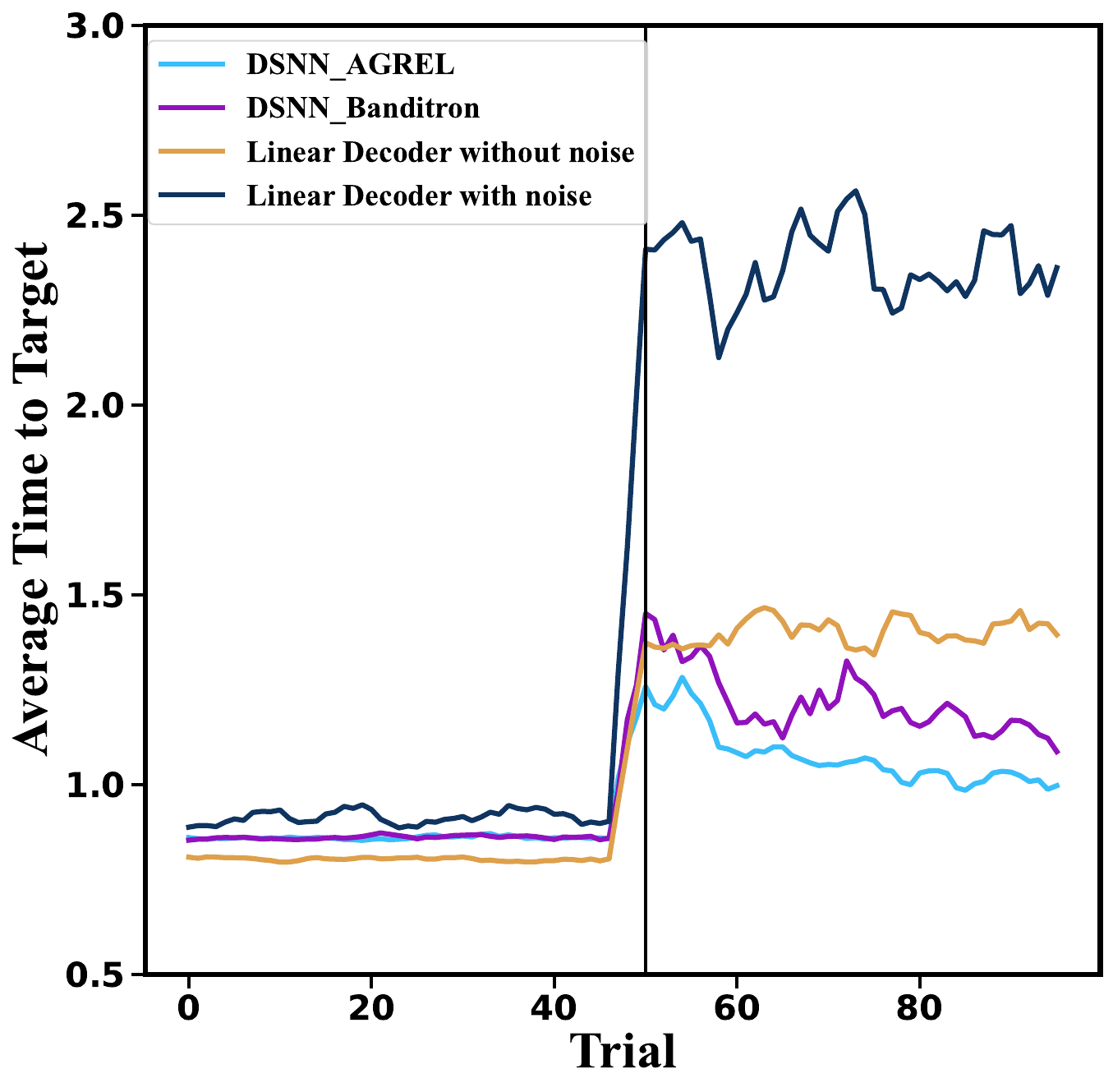}
    \caption{The closed-loop experiment on loss of neurons perturbation using linear decoder with/without a noisy brain model.}
    \label{fig:lineardecoder}
\end{framed}
\end{figure}

A linear decoder demonstrates remarkably strong performance, standing out among numerous advanced algorithms due to its remarkably low computational cost and straightforward architecture \cite{7123}. This observation prompts an inquiry: if a simple linear decoder is capable of solving this problem, is there still necessary to employ more complex methodologies?

The OPS environment described in \cite{7123} is ideal, as it assumes the brain signal is free from noise and the cursor can accurately reach a specified position. However, this scenario does not reflect the complexities inherent in real-world applications. To improve this process, we introduced noise into the brain model, characterized by a mean of 0 and a standard deviation of 0.3, consistent with the statement in \cref{subsec:closedloopexperiments}. We subsequently retrained the linear decoder utilizing this modified noisy brain model and repeated the closed-loop experiments while applying perturbations with the loss of neurons. The comparative results for the linear decoder, both with and without noise, are illustrated in \cref{fig:lineardecoder}. For comparison, the results of the continuous learning in this work are also included.

It is evident that the linear decoder performs poorly in a noisy environment, and its time to target increases significantly when the perturbations are applied. In contrast, the DSNN utilizing continuous learning strategies exhibits consistent and stable performance under similar conditions. These results indicate that a more complex decoder is necessary, as the basic linear decoder lacks sufficient abilities for effective implementation in practical, real-world scenarios.

\subsection{Future Directions}
\label{sec:futuredirections}

We implemented a closed-loop iBMI system that integrates an OPS-based brain model for real-time interaction. This system allows the OPS brain model to simulate brain signal generation. The decoder generates motor commands and receives a subsequent reward signal after the movement for self-calibration. The RL algorithms are integrated with the DSNN model to develop an energy-efficient continuous learning decoder that operates efficiently in both the forward and backward paths. This method effectively addresses the challenges associated with non-stationarity in the real-world iBMI systems, while also considering the energy constraints associated with the implantable device. However, from the perspective of this system, the brain model fundamentally lacks memory and produces new neural states at each time step based entirely on the input provided. Moreover, in this brain model, the representation of the intended direction within a 2D space may constrain its expressive potential. Therefore, the improvement of the brain model for the future of closed-loop systems should be approached from both perspectives.

From the perspective of the decoder, the exceptional energy efficiency of DSNNs is a result of the inherent sparsity observed during the forward processes. During inference, activity is sparse as only a small subset of neurons spike at any time. However, the computation involved in the backward process is dependent on the specific approach utilized. Algorithms related to Banditron exhibit performance levels comparable to those associated with AGREL; however, they require significantly lower computational resources, being 60 times less demanding than AGREL. This highlights the effectiveness and efficiency of transfer learning and Banditron in practical applications. In this study, we employed a binary feedback signal at each time step to provide update instructions to the model, as this form of signal can be generated by biological sources. Some RL algorithms aim to design a more effective guidance signal, such as actor-critic RL \cite{zhang2024kernel}. However, it is also necessary to account for the computational and energy requirements in these algorithms to evaluate their efficiency. Moreover, all models are stored employing 32-bit floating-point numbers, but in the practical application, quantization has to be implemented, which can be explored in the future. 

\section{Conclusion}
\label{sec:Conclusion}

The implementation of data compression in wireless iBMI systems is essential to meet the energy constraints and transmission budget. Integrating a neural decoder in the implant provides the maximum amount of compression. However, the design of the decoder needs to take into account the non-stationary characteristics of the iBMI system. The integration of DSNNs with RL algorithms and transfer learning establishes a highly promising framework for self-adaptive, low-power intention decoders for iBMI. DSNNs contribute significant energy efficiency through their event-driven and asynchronous processing capabilities. Meanwhile, the Banditron algorithm provides an effective, energy-efficient mechanism for continuous learning. The proposed method demonstrates the lowest computational requirements during the update process while achieving one of the highest levels of decoding accuracy in both open-loop and closed-loop experiments. Additionally, it also demonstrates superior time-to-target in the perturbation experiments conducted under closed-loop experiments.

\section*{Acknowledgment}
The work done in this paper was partially supported by a grant from the Research Grants Council of the Hong Kong Special Administrative Region, China (Project No. CityU 11200922).

\section{Appendix}
\label{sec:appendix}

\subsection{Analysis of the Computational Complexity of Continuous Learning Algorithms}
\label{sec:weightupdate}

List of notations used in this section
\begin{itemize}
    \item $MF_i$: The memory footprint in $i^{th}$ layer
    \item $MA_{i}^{forward}$: The number of memory access of $i^{th}$ layer in forward path
    \item $MA_{i}^{backward}$: The number of memory access of $i^{th}$ layer in backward path
    \item $N_{i}$: The number of neurons in $i^{th}$ layer
    \item $s_{i}$: The sparsity of $i^{th}$ layer
    \item $s_{e_{i}}$: The sparsity of error signal in $i^{th}$ layer
    \item $s_{fb_{i}}$: The sparsity of feedback signal in $i^{th}$ layer
    \item $AC_{i}^{forward}$: The number of accumulation in $i^{th}$ layer in forward path
    \item $MAC_{i}^{forward}$: The number of multiply-accumulate in $i^{th}$ layer in forward path
    \item $MAC_{i}^{backward}$: The number of multiply-accumulate in $i^{th}$ layer in backward path
    \item $T_s$: The number of time steps in SNNs
\end{itemize}

 \begin{enumerate}

    \item \textit{Baseline DSNN}

The memory footprint refers to the amount of memory required for models. The memory footprint calculations are detailed in \cref{eq:memory_footprint}, which uses a 32-bit floating number for each parameter.

\begin{align}
    MF_{i} = \begin{cases}
    ((N_{i}*N_{i-1})+2N_{i})*32 , & \text{if SNNs} \\
    ((N_{i}*N_{i-1})+N_{i})*32 , & \text{if ANNs} \\
    \end{cases}
\label{eq:memory_footprint}
\end{align}

where $N_{i}$ denotes the number of neurons in layer i. In this section, $N_0=46, N_1=65, N_2=40, N_3=8$, the numbers in \Cref{tab:computes} are calculated using these functions, while some of the numbers are derived from actual experiments. Furthermore, memory access is estimated based on the number of weight fetches, which is determined by experimentally observed sparsity multiplying the total number of weights \cref{eq:memory_access}. For simplicity, all the sparsity ($s_{i}$) in forward path used in this section for calculations is set to 0.6, since it is close to the sparsity levels observed in the experiments.

\begin{align}
    MA_{i}^{forward} = \begin{cases}
    (1-s_{i-1})*(N_{i}*N_{i-1})+2N_{i} , & \text{if SNNs} \\
    (1-s_{i-1})*(N_{i}*N_{i-1})+N_{i} , & \text{if ANNs} \\
    \end{cases}
\label{eq:memory_access}
\end{align}

In the baseline DSNN, there is no learning occurring on the backward path, so the number of computations is not considered. In the forward path, the number of ACs is significantly greater than MACs, because the ACs are mainly derived from binary neuron activation, while MACs primarily involves the calculation of the membrane potential.

\begin{align}
    AC_{i}^{forward} = \begin{cases}
    ((1-s_{i-1})*(N_{i}*N_{i-1})+2N_{i})*T_{s} , & \text{if SNNs} \\
    (1-s_{i-1})*(N_{i}*N_{i-1})+N_{i} , & \text{if ANNs} \\
    \end{cases}
\label{eq:SNN_forward_acs}
\end{align}

\begin{align}
    MAC_{i}^{forward} = \begin{cases}
    N_{i}*T_{s} , & \text{if SNNs} \\
    (1-s_{i-1})*(N_{i}*N_{i-1}) , & \text{if ANNs} \\
    \end{cases}
\label{eq:SNN_forward_macs}
\end{align}

where $T_{s}$ denotes the time steps used SNN, and in our case $T_{s}=1$. The \cref{eq:SNN_forward_acs}
 and \cref{eq:SNN_forward_macs} demonstrate the calculations of MAC and AC for ANN/SNN at the $i^{th}$ layer during the forward path. However, in the backward path, the computations differ among various algorithms, the analysis is presented as follows. 
 
 \item \textit{DSNN\_Banditron / Banditron}
 
 In the Banditron algorithm, only the weights of the last layer need to be updated; the number of memory access and computations is straightforward:

 \begin{align}
    MA^{Banditron} = 
    (1-s_{in}[t])*N_{in}*2 \nonumber \\
    MAC^{Banditron} = 
    (1-s_{in}[t])*N_{in}*2 
\label{eq:MAC_banditron}
\end{align}

where $N_{in}$ refers to the number of neurons in the previous layer. For Banditron, $N_{in}= N_{0}=46$ while $N_{in}= N_{2}=40$ for DSNN\_Banditron. The equation needs to be multiplied by 2, as the classification operates independently in both the x and y directions.

\item \textit{DSNN\_AGREL}

The update mechanism for DSNN\_AGREL is elaborated in \cref{subsec:continuouslearning}. The process for weights updating is divided into three components, as specified in \cref{eq:AGREL_update_layeri}: the calculation of the error signal, the computation of the feedback signal, and the update of the weight matrix. Thus, the total number of MAC is $MAC_{total} = MAC_{backward} + MAC_{feedback} + MAC_{error}$:

 \begin{align}
    MAC_{backward}^{i}[t] = 
    (1-s_{i-1}[t])*(1-s_{fb_{i}}[t])*N_{i-1}*N_{i}) \nonumber \\
    MAC_{feedback}^{i}[t] \approx 0 \nonumber \\
    MAC_{error}^{i}[t] =
    (1-s_{fb_{i+1}}[t])*N_{i+1}*N_{i}
\label{eq:MAC_AGREL}
\end{align}

Where $s_{i-1}[t]$ denotes the sparsity of spike output for layer $i-1$ at time t. Similarly, $s_{fb_{i}}[t]$ indicates the sparsity of feedback for layer $i$ at time t. For simplicity, the sparsity of feedback $s_{fb_{i}}$ is set to an average value of 0.94 in the backward path (obtained by running the source code). The MAC operation in the feedback is derived from the \cref{eq:AGREL_update_layeri}, which illustrates the multiplication of the error signal and the spike output of layer i. This value is approximately equal to zero because it does not require a MAC operation, only selecting ``important" signals based on the non-zero value of the spike. This serves as a ``gate," selecting which signals are significant for the feedback, only a few signals are deemed significant. Similar analysis for the memory access $MA_{total} = MA_{backward} + MA_{feedback} + MA_{error}$:

 \begin{align}
    MA_{backward}^{i}[t] = 
    (1-s_{i-1}[t])*(1-s_{fb_{i}}[t])*N_{i-1}*N_{i} \nonumber \\
    MA_{feedback}^{i}[t] = (1-s_{i}[t])*(1-s_{e_{i}}[t])*N_{i} \nonumber \\
    MA_{error}^{i}[t] = (1-s_{fb_{i+1}}[t])*N_{i+1}*N_{i}
\label{eq:MA_AGREL}
\end{align}

Both the number of reads and writes in the memory are considered in the calculation of the memory access \cref{eq:MA_AGREL}.

\item \textit{CLSNN \cite{taeckens2024spiking}} 

The continuous learning SNN presented in \cite{taeckens2024spiking} shares a similar structure with the DSNNs discussed in this paper, both incorporating two hidden layers. As a result, the analysis of its forward path follows a similar approach to that presented earlier. To analyze the backward path, 
the update mechanism can be written as:

 \begin{align}
    H_{i}[t+1] = \alpha * H_{i}[t] + (1-\alpha)*U_{i-1}[t] \nonumber \\
    G_{i}[t+1] = \beta * G_{i}[t] + (1-\beta)*H_{i}[t] \nonumber \\
    \Delta W_{i}[t] = -\eta * G_{i}[t] * \sum(Y_{i} - Y) \cdot R_{i}
\label{eq:CLSNN}
\end{align}

where $W_{i}$ and $R_{i}$ refer to the weight matrix, and readout matrix of the $i^{th}$ layer, respectively. The derivative of spiking output with respect to weight can be expressed using $G_{i}$ and $H_{i}$. Here, $\alpha$ and $\beta$ represent the decay rates of synaptic current and membrane potential. Additionally, $Y_{i}$ denotes the readout value of the $i^{th}$ layer, while $Y$ represents the corresponding target value. Besides, the $G_{i}$ also needs to be updated at every time step. Therefore, the number of MACs in the $i^{th}$ layer can be estimated as: $2*N_{i}*N_{i-1}$. This indicates that the estimated number of MACs is double the changed parameters, while the number of memory accesses is estimated as the number of changed parameters.

\item \textit{Other methods} 

The actor-critic RL \cite{7252} contains two models: the actor model and the critic model. In the forward path, the critic model remains inactive and is exclusively utilized during the training phase. Similarly, in the backward path, the number of changed parameters in the source code is assessed by comparing the weight matrix before and after the learning process. The number of MACs is estimated to be twice the number of changed parameters, while memory access is estimated to be the number of changed parameters.

In the forward path of the EGRU model \cite{evnn2023}, the analysis of the linear layer follows the same procedure as above. The computations associated with the EGRU cell are estimated as: $3*[N_{h}*(N_{h}+N_{in})+2N_{h}]$. Where $N_{h}$ represents the size of the hidden state, while $N_{in}$ denotes the number of input for EGRU cell. In the backward pass of the EGRU cell, the computation of the error approximately involves two multiplications and one matrix multiplication, while gradient calculation involves one matrix multiplication. Therefore, the total update process in the EGRU cell is estimated to be four times the number of changed parameters. The number of memory access is estimated as twice the number of changed parameters, since the calculation involves the hidden state of the previous time step.

 \end{enumerate}

\clearpage

\section*{References}
\bibliography{refs.bib}
\bibliographystyle{IEEEtran}

\end{document}